# Polarized skylight orientation determination artificial neural network

Huaju Liang,[1] Hongyang Bai,[1,*] Ke Hu,[1] and Xinbo Lv[2]

[1]*School of Energy and Power Engineering, Nanjing University of Science and Technology, Nanjing 210094, China*
[2]*School of Automation, Nanjing University of Science and Technology, Nanjing 210094, China*
*\*hongyang@mail.njust.edu.cn*

**Abstract:** This paper proposes an artificial neural network to determine orientation using polarized skylight. This neural network has specific dilated convolution, which can extract light intensity information of different polarization directions. Then, the degree of polarization (DOP) and angle of polarization (AOP) are directly extracted in the network. In addition, the exponential function encoding of orientation is designed as the network output, which can better reflect the insect's encoding of polarization information, and improve the accuracy of orientation determination. Finally, training and testing were conducted on a public polarized skylight navigation dataset, and the experimental results proved the stability and effectiveness of the network.

## 1. Introduction

It has been found that the orientation behaviours of many insects are based on skylight polarization patterns, such as desert ants, locusts, dung beetles, and so on [1]. In the dorsal rim area (DRA) of the insect compound eye, microvilli are neatly arranged in different directions which can perceive the light intensity information in different polarization directions [2, 3]. Then, to process the light intensity information, polarized skylight signals are integrated in a group of neuropils in the center of the brain [4, 5]. Finally, compass neurons respond differently when the insect orient in different directions, to get a specific orientation encoding and determine orientation [6-9].

Inspired by insects, a variety of orientation determination methods using skylight polarization patterns have been proposed, such as: zenith method (polarization electric-field vectors (E-vectors) at zenith are perpendicular to the solar azimuth) [10-12], solar meridian methods (polarization E-vectors at solar meridian and anti-solar meridian (SM-ASM) are perpendicular to the solar azimuth) [13, 14], symmetry method (skylight polarization patterns are symmetric about SM-ASM) [15, 16], least square method (polarization E-vectors are perpendicular to the solar vector) [17, 18]. These traditional methods are based on a certain characteristic of skylight polarization patterns to determine orientation, so skylight polarization patterns are not fully utilized by these traditional methods. When the used characteristics are disturbed, the error of orientation determination increases. Therefore, artificial neural networks are applied to polarized skylight orientation determination.

Artificial neural networks are widely used in the field of image processing, and some scholars have begun to use artificial neural networks to process polarization images and extract polarization information [19-22]. These methods are mainly aimed at target recognition and noise reduction, and have achieved certain results in the field of image processing. However, these networks are complex and not for polarized skylight navigation, so it is difficult to apply them directly in the field of polarized skylight orientation determination.

The artificial neural networks applied to polarized skylight orientation determination can be divided into given parameter neural networks [23] and training parameter neural networks [24]. Given parameter neural networks only express orientation determination in the form of a neural network, which is essentially the same as the traditional method. The training parameter neural network has been trained to obtain network parameters. However, it is only a simple fully connected network, and the output of this network is orientation directly, which is difficult to reflect the encoding of orientation information by insects and resulting in unstable orientation determination.

In order to ensure the effective use of polarization information and reflect the encoding of insect neural network system, this paper proposes an artificial neural network to determine orientation, whose special neural network structure can directly extract the degree of polarization (DOP) and angle of polarization (AOP) directly from the original light intensity information. And inspired by the insect's orientation encoding, we designed the exponential function encoding of orientation as the network output to improve the accuracy and reliability of orientation determination.

## 2. Methods

As shown in Fig. 1, the photoreceptors of the dorsal rim area (DRA) of the compound eye are polarization sensitive and can detect light intensity information in different polarization directions [2, 5]. Then, these information is integrated by the insect polarization nervous system, as shown in Fig. 1(d) [4, 6]. Finally, they are transmitted to many compass neurons. As shown in Fig. 1(e), these compass neurons have different responses to body orientation of the insect, and each respond maximally when the insect is oriented in a particular compass direction. As show in Fig. 1(f), these compass neurons together constitute the orientation encoding, which makes insects can determine orientation stably and reliably [6, 9].

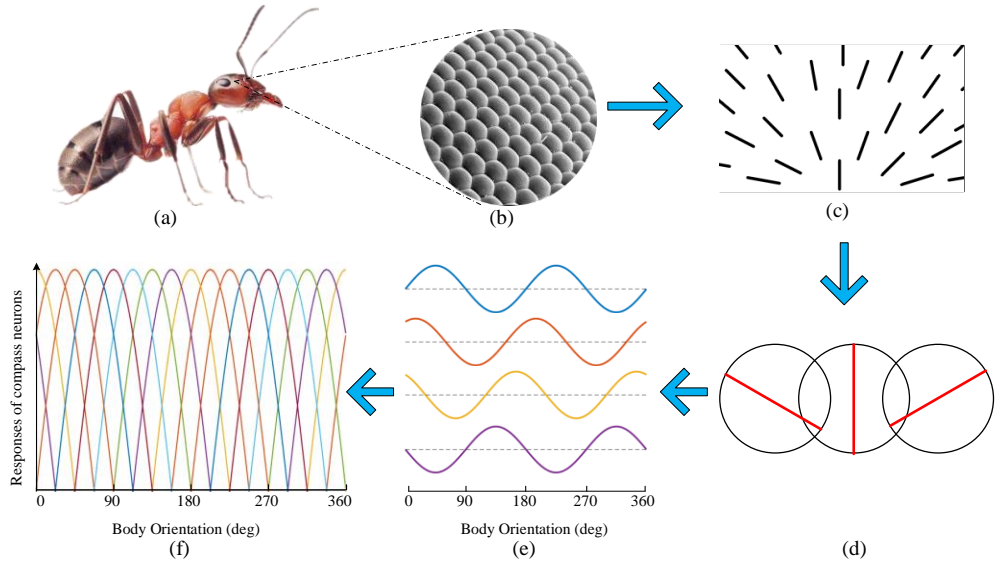

Fig. 1. Insect polarized skylight orientation determination: (a) Insect; (b) Compound eye; (c) Polarization-sensitive photoreceptors; (d) Integrators; (e) Compass neurons; (f) Orientation encoding.

Inspired by insects, the architecture of polarized skylight orientation determination artificial neural network is shown in Fig. 2.

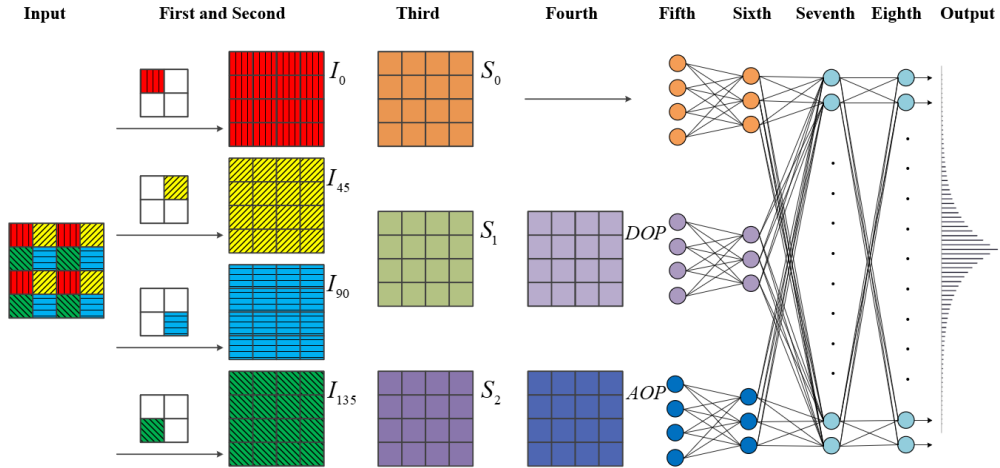

Fig. 2. architecture of polarized skylight orientation determination artificial neural network, where $I_0$, $I_{45}$, $I_{90}$ and $I_{135}$ represent light intensity information of 0°, 45°, 90° and 135° polarization directions. $(S_0 \ S_1 \ S_2 \ S_3)$ represents Stokes vector. DOP represents the degree of polarization, and AOP represents the angle of polarization.

As shown in Fig. 2, this artificial neural network consists of the following layers
Input layer: Original intensity image；
First layer: Specific dilated convolution；
Second layer: Mean-pooling；（Not shown in Fig.2）
Third layer: Stokes vector；
Fourth layer: DOP and AOP；
Fifth layer: Locally full connection；
Sixth layer: Locally full connection；
Seventh layer: Globally full connection；
Eighth layer: Globally full connection；
Output layer: Orientation encoding;
And, this artificial neural network can be divided into three parts: original light intensity information processing part (from the Input layer to the Fourth layer), feature extraction part (from the Fifth layer to the Eighth layer), and encoding output part (Output layer).

## 2.1 Original light intensity information processing

The neural network directly takes the original light intensity image as the input, and the first four layers of the network complete the processing of the original light intensity information.

In order to obtain the polarization information, the light intensity information of different polarization directions is collected. Sony PHX050S-P pixel polarization camera can capture light intensity information of four polarization directions (Fig. 2 Input layer) in one picture at a shot, and it has a small size and real-time performance. Therefore, the camera can detect polarized light just like the compound eye of insects as shown in Fig.1 (a) and Fig.1 (b). The network input is directly the original light intensity image collected by this polarization camera. As shown in Fig. 3 and Fig.2 First layer, through the specific dilated convolution with a step length of 2, the four-channel original light intensity images are extracted, which are the light intensity information of different polarization directions.

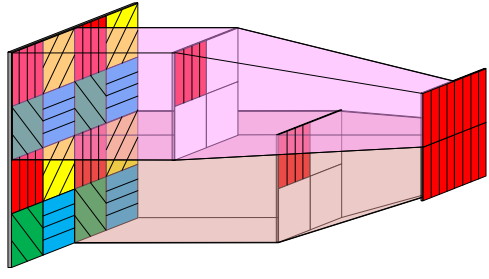

Fig. 3. Specific dilated convolution.

The mean-pooling (mean filter) is used to compress the light intensity images of different polarization directions, which can not only reduce the influence of random noise, but also reduce the amount of calculation. Its function is similar to the integrator in Fig.1 (d).

Next, for the compressed light intensity images, the Stokes vector $(S_0\ S_1\ S_2\ S_3)$ is directly calculated in the network to describe the polarization state. Circularly polarized light in the sky is very rare, so $S_3 = 0$. Therefore, the Stokes vector is calculated as follows

$$\begin{cases} S_0 = \frac{1}{2}[I_0 + I_{45} + I_{90} + I_{135}] \\ S_1 = I_0 - I_{90} \\ S_2 = I_{45} - I_{135} \\ S_3 = 0 \end{cases} \quad (1)$$

where $I_0$, $I_{45}$, $I_{90}$ and $I_{135}$ represent light intensity information of $0°$, $45°$, $90°$ and $135°$ polarization directions. And to reduce the amount of calculation, directly ignore $S_3$, as shown in Fig. 2 in the Third layer to get the Stokes vector graph $S_0$, $S_1$, $S_2$.

The fourth layer of the network completes the extraction of DOP and AOP

$$DOP = \frac{\sqrt{S_1^2 + S_2^2}}{S_0} \quad (2)$$

$$AOP = \frac{1}{2}\arctan\left(\frac{S_2}{S_1}\right) \quad (3)$$

Above all, in this part, a special convolution structure is designed to process the original light intensity information, which can directly extract DOP and AOP. Therefore, it omits the separate process of converting the light intensity information into the DOP and AOP, and the graphics processing unit (GPU) can be used for this calculation, the processing speed is greatly improved.

*2.2 Feature extraction*

The feature extraction part consists of a local full connection network (Fifth layer and Sixth layer) and a global full connection network (Seventh layer and Eighth layer), whose activation function is Sigmoid function. Firstly, two layers of local full connection are used to extract the characteristics of total light intensity ($S_0$), DOP, and AOP. This is because $S_0$, DOP and AOP are different types of information, and their characteristics are not the same, so the local full connection distribution is used to extract their own characteristics.

After each feature is extracted, the feature information is collected and fused through global full connection, and make full use of the information of $S_0$, DOP and AOP contained in skylight polarization patterns.

*2.3 Encoding output*

As shown in Fig. 4, the network output design has gone through the following process: 0-360 encoding, 0-1 encoding, One-Hot encoding, trigonometric function encoding, exponential function encoding. The design idea will be described in detail below.

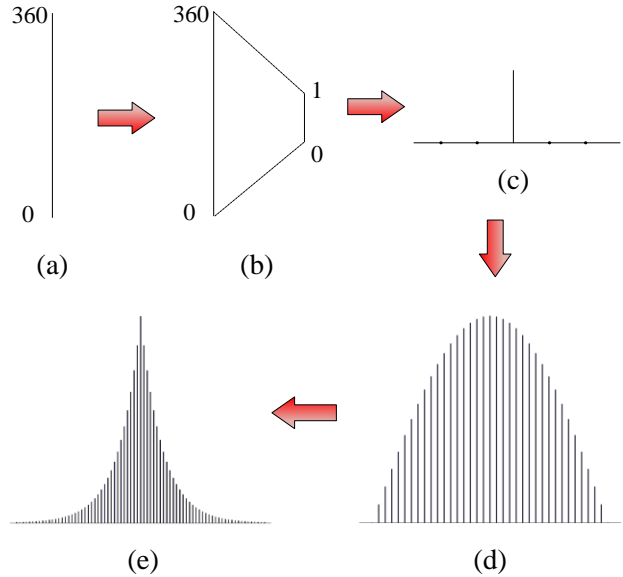

Fig. 4. Encoding output: (a) 0-360 encoding; (b) 0-1 encoding; (c) One-Hot encoding; (d) trigonometric function encoding; (e) exponential function encoding.

Because the purpose of the network is to determine orientation, it is easy to think of the network output directly the orientation from 0 degrees to 360 degrees, that is, the network output is 0-360 [24]. But the fatal problem of this output is that the output is not normalized, which makes the network difficult to train, and even the network does not converge.

Therefore, we consider normalizing the orientation from 0-360 to 0-1 as the network output, so that 0 corresponds to 0 degrees, 0.5 corresponds to 180 degrees, 1 corresponds to 360 degrees.

But the output normalized to 0-1 is still problematic. 0 degrees and 360 degrees are equivalent, but the output of 0-360 encoding and the normalized 0-1 encoding output will not only make the output of 0 degrees and 360 degrees not equivalent, but also have the largest difference, which is obviously unreasonable.

Therefore, further, we consider using One-Hot encoding as network output. It is a process that converts data to a vector of 1 and 0, and only one bit of the vector is 1. Due to One-Hot encoding's effectiveness and simplicity, it is the most prevalent procedure for addressing many multi-class classification tasks [25]. Using One-Hot encoding as the network output can ensure that the output of the network corresponding to different heading angles is equivalent. This solves the problem that 0 degrees and 360 degrees are not equivalent and have the largest difference.

However, One-Hot encoding does not reflect that the angular distance from 0 degrees to 10 degrees is less than that from 0 degrees to 20 degrees, nor does it reflect the correlation between neurons. In addition, only relying on the output of a single neuron will cause the entire network to be easily disturbed.

Inspired by the response of insect compass neurons to polarized light with different polarization directions, as shown in Fig.1 (e) and Fig.1 (f), consider using trigonometric function encoding to make multiple neurons related to each other to improve the stability of the network and reflect that the angular distance from 0 degrees to 10 degrees is less than that from 0 degrees to 20 degrees.

$k$ represents the $k$ th neuron. For trigonometric function encoding, the output of $k$ th neuron is

$$N_t(k) = \cos(i \times j) \quad (0 \leq k < \frac{360}{j} \text{ and } k \text{ is an integer}) \quad (4)$$

where

$$k = i + \frac{\varphi}{j} \quad (5)$$

where $\varphi$ is orientation, $j$ is the angular resolution of encoding. $-\frac{90}{j} < i < \frac{90}{j}$ and $i$ is an integer, which ensures that the value of trigonometric function encoding is greater than 0. And considering that 0 degrees and 360 degrees are equivalent. If $k \geq \frac{360}{j}$, $k = k - \frac{360}{j}$. If $k < 0$, $k = k + \frac{360}{j}$. Each neuron corresponds to an orientation, and the $k$ th neuron corresponds to $k \times j$ degrees orientation.

We take $j=0.1°$, $\varphi=26°$ as an example to illustrate the trigonometric function encoding. At this time, the 260th neuron corresponds to 26 degrees. $k_{26}^{26} = 0 + \frac{26}{0.1} = 260$, so $i=0$. The superscript of $k_{26}^{26}$ indicates the true value of the orientation, and the subscript of $k_{26}^{26}$ indicates the angle corresponding to the current neuron. The neuron output corresponding to 26 degrees is $N_t(k_{26}^{26}) = \cos(0 \times 0.1) = 1$. At this time, the neuron output corresponding to 26.1 degrees is $N_t(k_{26.1}^{26} = 261) = \cos(1 \times 0.1) = \cos(0.1)$. At this time, the neuron output corresponding to 25.9 degrees is $N_t(k_{25.9}^{26} = 259) = \cos(-1 \times 0.1) = \cos(0.1) = N_t(k_{26.1}^{26})$, which reflects the angular distance from 26.1 degrees to 26 degrees is the same as the angular distance from 25.9 degrees to 26 degrees. 0 degrees and 360 degrees correspond to the same neuron and have the same output $N_t(k_0^{26} = 0) = N_t(k_{360}^{26} = 0) = \cos(26)$, which avoids the problem of the smallest angular distance but the largest encoding distance in 0-360 encoding and 0-1 encoding. In addition, $\cos(26) < \cos(0.1)$. This shows the activation level of neurons at 26.1 (25.9) degrees is greater than the activation level of neurons at 0 (360) degrees, which reflects the angular distance from 26.1 (25.9) degrees to 26 degrees is less than the angular distance from 0 (360) degrees to 26 degrees. Moreover, trigonometric function encoding makes multiple neurons related to each other at an angle, which improves the stability of the network.

But the trigonometric function encoding has a major flaw. The output of the neuron corresponding to the true orientation differs very little from the output of the neighboring neuron. Taking $j=0.1°$ and $\varphi=26°$ as an example, the neighboring neurons are the neurons corresponding to 26.1 degrees and 25.9 degrees, and their output is $\cos(0.1) = 0.99999847691329$. So, the output difference between 26 degrees neuron and neighboring neurons is $1-\cos(0.1) \approx 0.00000152$. It can be found that the difference is very small, which will result in that 26 degrees orientation is very likely to be calculated as 26.1 degrees or 25.9 degrees. This means that the orientation determination is not accurate enough.

Therefore, consider using exponential function encoding as the network output. $k$ represents the $k$ th neuron. For exponential function encoding, the output of $k$ th neuron is

$$N_e(k) = m^{|i|} \quad (0<m<1,\ 0\leq k < \frac{360}{j}\ \text{and}\ k\ \text{is an integer})\quad (6)$$

where

$$k = i + \frac{\varphi}{j} \quad (7)$$

$m$ is a hyperparameter, $j$ is the angular resolution of encoding, $\varphi$ is orientation and $i$ is an integer. Considering that 0 degrees and 360 degrees are equivalent. If $k \geq \frac{360}{j}$, $k=k-\frac{360}{j}$. If $k<0$, $k=k+\frac{360}{j}$. Each neuron corresponds to an orientation, and the $k$ th neuron corresponds to $k \times j$ degrees orientation.

When $m=0.98$, continue to take $j=0.1°$ and $\varphi=26°$ as an example, the neuron output corresponding to 26 degrees is 1 ($N_e(k_{26}^{26} = 260) = 0.98^0 = 1$), and the neuron outputs corresponding to 26.1 and 25.9 degrees is 0.98 ($N_e(k_{26.1}^{26} = 261) = 0.98^1 = 0.98 = N_e(k_{25.9}^{26} = 259)$). The output difference of neurons is $1-0.98 = 0.02 \gg 1-\cos(0.1)$.

It can be seen that exponential function encoding not only has the advantage of trigonometric function encoding but also the output of the true value neuron and the neighboring neuron has a certain difference, so that the orientation can be calculated with higher accuracy.

According to exponential function encoding, define loss function as below:

$$Loss = \sum_{k=0}^{\frac{360}{j}} \left(N_e^n(k) - N_e^t(k)\right)^2 \quad (k\ \text{is an integer})\quad (8)$$

where $N_e^n(k)$ is the exponential function encoding output by the network, $N_e^t(k)$ is exponential function encoding of true orientation. It is obvious that the loss function correlates all the output neurons, which makes the output of the network more stable.

## 3. Experiments

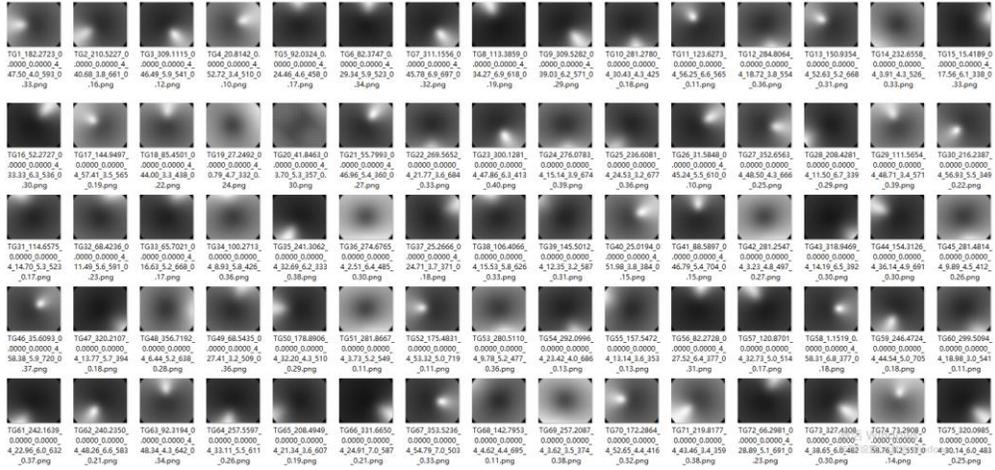

Fig. 5. Original light intensity images in TGNIIPI folder of PSNS dataset.

In order to verify the effectiveness of the designed network, we train and test on the public polarized skylight navigation simulation (PSNS) dataset [26]. This article mainly focuses on the use of neural network to determine orientation in sunny weather, so the following uses about 36000 original light intensity images in GNIIPI folder of PSNS dataset as the training set, and about 10000 original light intensity images in TGNIIPI folder of PSNS dataset as the test set. Fig. 5 shows the images in TGNIIPI folder of PSNS dataset.

*3.1 Encoding comparison*

We compare the 0-360 encoding proposed in reference [24] with the other four encoding (0-1 encoding, One-Hot encoding, trigonometric function encoding, exponential function encoding) proposed in this paper. In the experiment, the angular resolution of One-Hot encoding, trigonometric function encoding, exponential function encoding is 0.1 degrees ($j=0.1°$), and hyperparameter $m$ of exponential function encoding is 0.98 ($m=0.98$). We trained 60 batches on the GNIIPI folder of PSNS dataset, and then tested on TGNIIPI folder of PSNS dataset. The orientation determination results are shown in Fig.6 and Table 1. Note that the orientation determination result in Fig.6 and Table 1 has been processed. If the orientation determination error is greater than 90 degrees, the error is subtracted by 180 degrees. The purpose of this is to see the error comparison when the orientation determination range is 0-180 degrees. Orientation determination in the range of 0-360 degrees will be described in section 3.2.

**Table 1. Encoding comparison**

| Encoding | 0-360 | 0-1 | One-Hot | Trigonometric | Exponential |
|---|---|---|---|---|---|
| MAE (deg) | 54.688 | 18.283 | 22.603 | 1.220 | 0.337 |
| ME (deg) | 89.999 | 89.656 | 89.994 | 39.264 | 4.424 |
| RMSE (deg) | 60.715 | 22.958 | 33.056 | 1.666 | 0.460 |

mean absolute error (MAE), root mean square error (RMSE), maximum error (ME), degree (deg)

It is obvious that the mean absolute error (MAE), root mean square error (RMSE) and maximum error (ME) of 0-360 encoding are the largest among the five methods, which may be due to the non-normalization of the output and the inequivalence of 0 degrees and 360 degrees. MAE and RMSE of 0-1 encoding are less than that of 0-360 encoding. And as shown in Fig. 6(b), the error of 0-1 encoding is mainly below 45 degrees. These may be due to the normalized output. What needs special attention is the orientation determined by 0-360 encoding and 0-1 encoding is not between 0 degrees and 360 degrees. For example, an orientation determined by 0-360 encoding is 402.756 degrees, and an orientation determined by 0-1 encoding is 405.695 degrees, which have not physical meaning.

Compared with 0-1 encoding and 0-360 encoding, the advantages of One-Hot encoding, trigonometric function encoding, and exponential function encoding are highlighted, which can ensure that the calculated orientation is between 0 degrees and 360 degrees, which has physical meaning.

However, the orientation determination error of One-Hot encoding is still very large. One-Hot encoding only activates one neuron as the output of the network, so the network is unstable, vulnerable to interference. In addition, it cannot reflect the angular distance. Trigonometric function encoding can activate multiple neurons as network output, and can reflect the distance relationship of angle. Compared with One-Hot encoding, the orientation determination error of trigonometric function encoding is greatly reduced, and MAE and RMSE of trigonometric function encoding are only 1.220 degrees and 1.666 degrees, and the ME of trigonometric function encoding is also reduced to 39.264 degrees. However, because the output difference

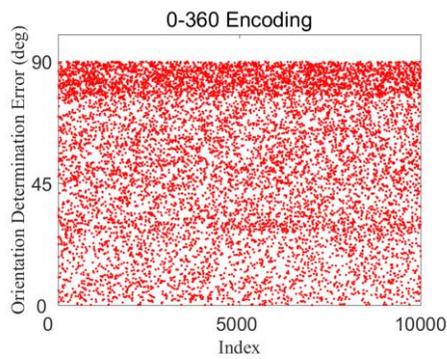

(a)

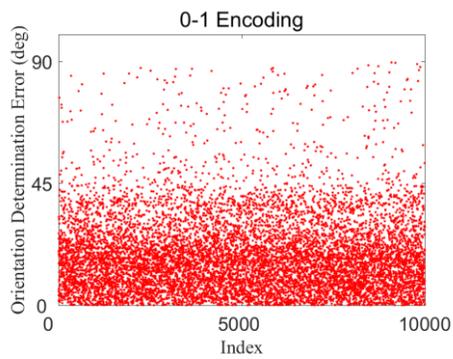

(b)

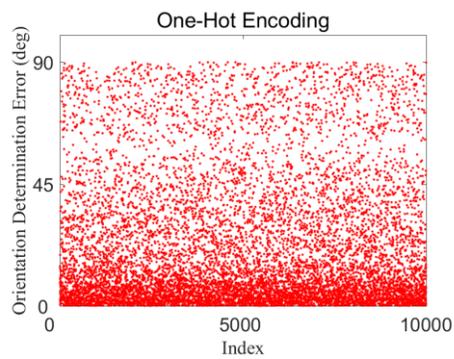

(c)

Fig. 6. Encoding comparison: (a) Orientation determination error of 0-360 encoding; (b) Orientation determination error of 0-1 encoding; (c) Orientation determination error of One-Hot encoding; (d) Orientation determination error of trigonometric function encoding; (e) Enlarged view of orientation determination error of trigonometric function encoding; (f) Orientation determination error of trigonometric function encoding; (g) Enlarged view of orientation determination error of trigonometric function encoding.

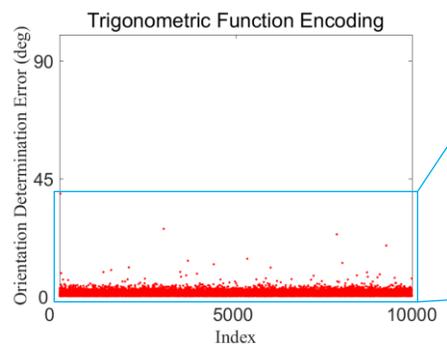

(d)

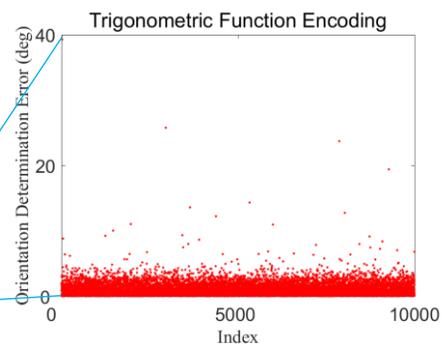

(e)

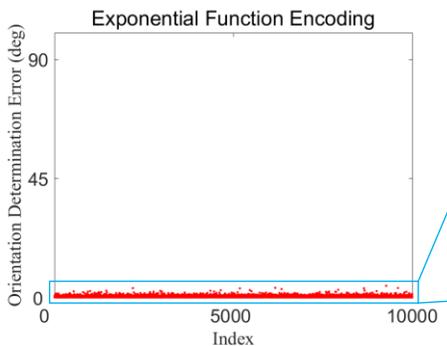

(f)

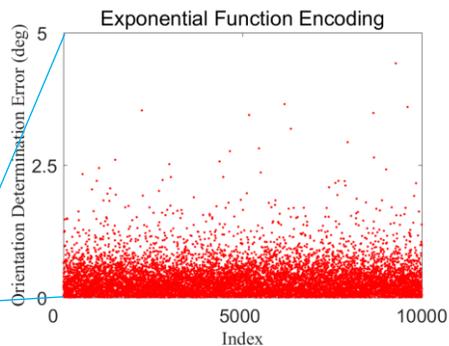

(g)

between the neuron corresponding to the true orientation and the neighboring neurons is very small, the accuracy of this method is still limited. Exponential function encoding is a good solution to this problem, as shown in Table 1, Fig. 6(f) and Fig. 6(g), the orientation determined by exponential function encoding has the smallest MAE (0.337 degrees), smallest RMSE (0.460 degrees) and smallest ME (4.424 degrees). And it takes only 0.044 seconds to complete an orientation determination on a GTX 1080Ti.

## 3.2 Explanation of random phenomena

The following will be explained for the orientation determination in the range of 0-360 degrees. We have done many experiments on exponential function encoding ($j=0.1$, $m=0.98$). The exponential function encoding artificial network has been trained and tested many times, and a random phenomenon has been found. The experimental results are shown in Fig. 7 and Table 2.

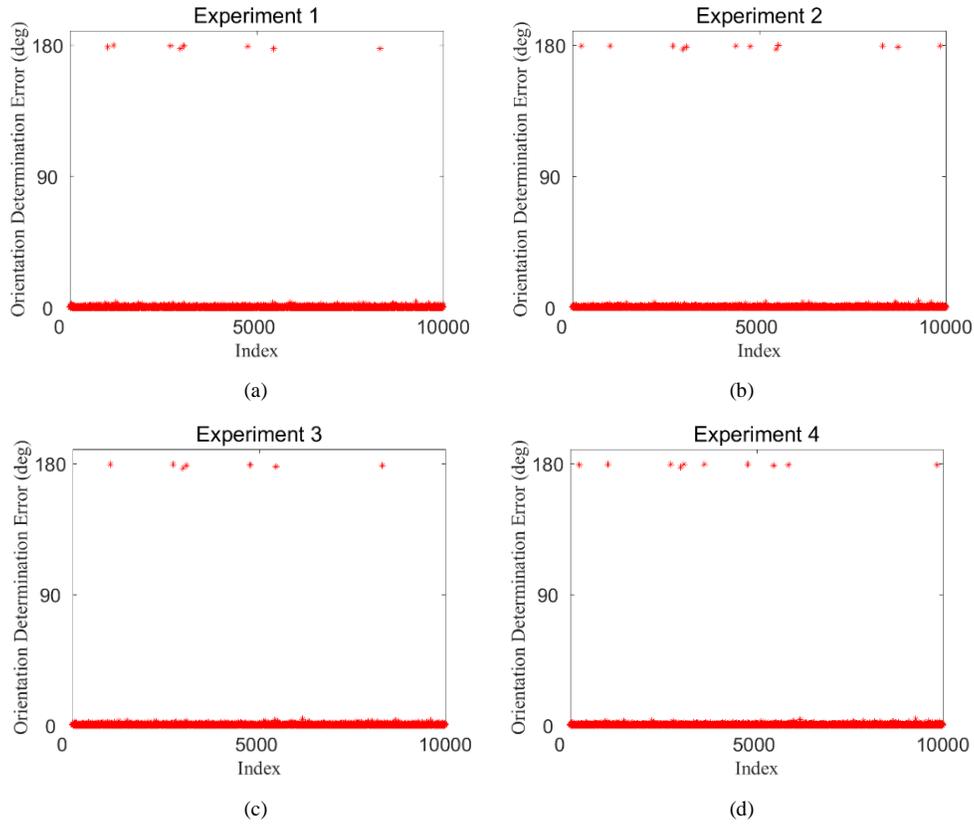

Fig. 7. 180 degree error random phenomena: (a) Experiment 1; (b) Experiment 2; (c) Experiment 3; (d) Experiment 4.

**Table 2. 180 degree error statistics**

| Experiments | Experiment 1 | Experiment 2 | Experiment 3 | Experiment 4 |
| --- | --- | --- | --- | --- |
| N180E | 8 | 12 | 7 | 10 |
| MSA (deg) | 0.55 | 0.95 | 0.55 | 0.43 |

number of 180 degree errors (N180E), maximum solar altitude (MSA), degree (deg)

It can be seen that in the four experiments, either the orientation determination error is about 180 degrees, or the error is less than 4.5 degrees. We extract these orientations with errors of about 180 degrees for observation, which can be seen in Table S1, Table S2, Table S3 and Table S4 for details. The statistical results are shown in Table 2. The following phenomena are found

1. The 180 degrees error has certain randomness. In different experiments, the number and position of 180 degrees error have a certain difference. For example, the four experiments have the different numbers of 180 degree errors. And as shown in Table S1, Table S2, Table S3 and Table S4, when we determine orientation on TG1217 of PSNS dataset, the orientation determination errors of Experiment 2 and Experiment 4 are about 180 degrees, but the orientation determination errors of Experiment 1 and Experiment 3 are less than 4.5 degrees.

2. The 180 degree error occurs when the solar altitude angle is very small. As shown in Table 2, the maximum solar altitudes are all less than 1 degrees, when 180 degree error occurs.

Skylight polarization patterns have a certain symmetry about the plane of 90 degrees away from the sun. When solar altitude is 0 degrees, the plane symmetry is more obvious. So, when solar altitude is 0 degrees, it is difficult to determine orientation in the range of 0-360 degrees, so orientation determination has error about 180 degrees. Therefore, when the solar altitude angle is very small, it is not recommended to determine orientation in the range of 0-360 degrees.

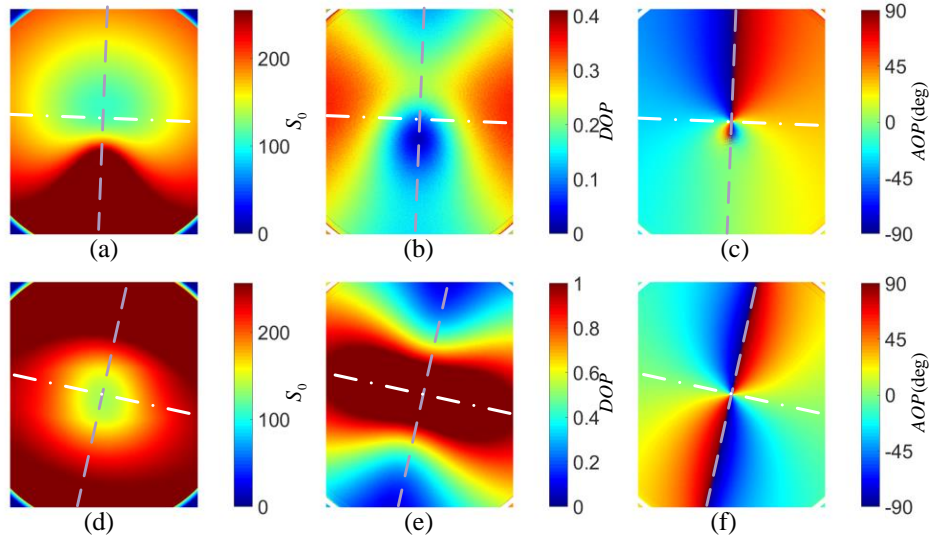

Fig. 8. (a) Light intensity ($S_0$) image of TG1 of PSNS dataset; (b) DOP image of TG1 of PSNS dataset; (c) AOP image of TG1 of PSNS dataset; (d) Light intensity ($S_0$) image of TG1906 of PSNS dataset; (e) DOP image of TG1906 of PSNS dataset; (f) AOP image of TG1906 of PSNS dataset. The dotted grey line is solar meridian and anti-solar meridian (SM-ASM). The white dot-and-dash line is the line which passes through the zenith and is perpendicular to SM-ASM.

In view of the above phenomenon, this section mainly answers two questions: (1) Why the neural network proposed in this paper can determine orientation in the range of 0-360 degrees in most cases. (2) Why the orientation can only be determined in the range of 0-180 degrees, when solar altitude angle is very small.

For (1), as shown in Fig.8 (a), Fig.8 (b) and Fig.8 (c), skylight polarization pattern is symmetric about SM-ASM, but However, skylight polarization pattern is not symmetrical about the line which passes through the zenith and is perpendicular to SM-ASM. Although the traditional method can extract SM-ASM, but they didn't take advantage of the asymmetry mentioned earlier. Therefore, they can almost only determine orientation in the range of 0-180 degrees. The neural network proposed in this paper is likely to extract the asymmetry, so it can determine orientation in the range of 0-360 degrees.

For (2), when the solar altitude angle is very small, as shown in t Fig.8 (d), Fig.8 (e) and Fig.8 (f), it can be found that skylight polarization pattern is almost symmetrical about the line which passes through the zenith and is perpendicular to SM-ASM. So, the orientation can only be determined in the range of 0-180 degrees.

*3.3 Hyperparameter Selection*

The third set of experiments will illustrate the choice of hyperparameters. We can find that there is an extra hyperparameter $m$ for exponential function encoding. In order to select the optimal hyperparameter $m$, training and testing were performed when $m=0.95$, 0.96, 0.97, 0.98, and 0.99. The experimental results are shown in Fig. 9.

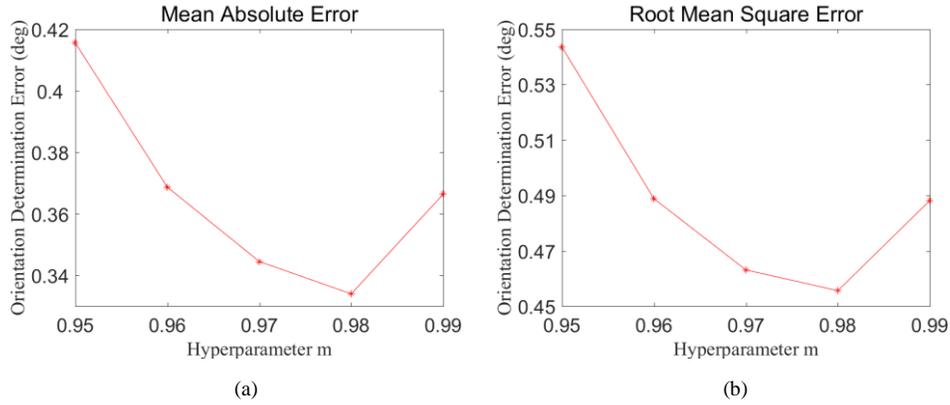

Fig. 9. Hyperparameter $m$ selection: (a)Mean absolute error of Hyperparameter selection; (b)Root mean square error of Hyperparameter selection.

When $m=0.98$, ME and RMSE of the heading angle solution are the smallest, so the hyperparameter m is set to 0.98. The reason for the hyperparameter to be 0.98 may be as follows. First, it is necessary to ensure that there is a certain output difference between the true value neuron and the neighboring neurons. When $m=0.99$, the difference is too small. At the same time, it is necessary to ensure that neighboring neurons are activated enough, so as to ensure the stability and reliability of the network. when $m=0.95$, 0.96, and 0.97, the neighboring neurons are not enough activated, so the stability of the network is not enough. When $m=0.98$, the difference between neighboring neurons can be guaranteed, and the neighboring neurons can be enough activated.

## 4. Conclusion

A polarized skylight orientation determination artificial neural network is designed in this paper. This paper has two major contributions. 1. A special dilated convolution structure is designed to process the original light intensity information, which can directly extract DOP and AOP in the artificial neural network. 2. Inspired by insects, orientation encoding is designed as the network output, which can better reflect the insect's encoding of the polarization information, and improve the accuracy and reliability of orientation determination. Finally, the experimental results also prove the effectiveness of the network.

This paper mainly focuses on orientation determination under clear weather. So, we will further improve the network to determine orientation under cloudy weather. Biological inspiration has a great role in promoting the development of technology. We will continue to study biological-related neural structures and apply them to polarized skylight navigation.

**Funding**

**Disclosures**

The authors declare no conflicts of interest.

## Supplementary materials

**Table S1. 180 degree error of Experiment 1**

| Index | Orientation(deg) | Determined Orientation(deg) | Solar Altitude(deg) | Orientation Error(deg) |
|---|---|---|---|---|
| TG1906 | 12.6307 | 191.4 | 0.09 | 178.7693 |
| TG2053 | 336.4890 | 156.4 | 0.10 | 179.9110 |
| TG3427 | 269.1191 | 89.4 | 0.06 | 179.7191 |
| TG365 | 124.9191 | 302.5 | 0.17 | 177.5809 |
| TG373 | 1.1286 | 181.6 | 0.43 | 179.5286 |
| TG5289 | 277.0299 | 97.7 | 0.03 | 179.3299 |
| TG5911 | 286.8179 | 104.4 | 0.31 | 177.5821 |
| TG8484 | 4.9218 | 187.3 | 0.55 | 177.6218 |

**Table S2. 180 degree error of Experiment 2**

| Index | Orientation(deg) | Determined Orientation(deg) | Solar Altitude(deg) | Orientation Error(deg) |
|---|---|---|---|---|
| TG1217 | 256.3126 | 76.7 | 0.15 | 179.6126 |
| TG1906 | 12.6307 | 192.4 | 0.09 | 179.7693 |
| TG3427 | 269.1191 | 89.5 | 0.06 | 179.6191 |
| TG365 | 124.9191 | 302.4 | 0.17 | 177.4809 |
| TG373 | 1.1286 | 182.3 | 0.43 | 178.8286 |
| TG4939 | 126.5618 | 306.2 | 0.08 | 179.6382 |
| TG5289 | 277.0299 | 97.9 | 0.03 | 179.1299 |
| TG5911 | 286.8179 | 104.0 | 0.31 | 177.1821 |
| TG5969 | 5.4063 | 185.4 | 0.95 | 179.9937 |
| TG8484 | 4.9218 | 185.4 | 0.55 | 179.5218 |
| TG8852 | 347.2467 | 168.2 | 0.02 | 179.0467 |
| TG9856 | 215.5961 | 35.3 | 0.08 | 179.7039 |

**Table S3. 180 degree error of Experiment 3**

| Index | Orientation(deg) | Determined Orientation(deg) | Solar Altitude(deg) | Orientation Error(deg) |
|---|---|---|---|---|
| TG1906 | 12.6307 | 192.1 | 0.09 | 179.4693 |
| TG3427 | 269.1191 | 88.6 | 0.06 | 179.4809 |
| TG365 | 124.9191 | 301.8 | 0.17 | 176.8809 |
| TG373 | 1.1286 | 182.2 | 0.43 | 178.9286 |
| TG5289 | 277.0299 | 97.8 | 0.03 | 179.2299 |
| TG5911 | 286.8179 | 104.9 | 0.31 | 178.0821 |
| TG8484 | 4.9218 | 186.0 | 0.55 | 178.9218 |

**Table S4. 180 degree error of Experiment 4**

| Index | Orientation(deg) | Determined Orientation(deg) | Solar Altitude(deg) | Orientation Error(deg) |
|---|---|---|---|---|
| TG1217 | 256.3126 | 77.0 | 0.15 | 179.3126 |
| TG1906 | 12.6307 | 192.2 | 0.09 | 179.5693 |
| TG3427 | 269.1191 | 88.7 | 0.06 | 179.5809 |
| TG365 | 124.9191 | 302.5 | 0.17 | 177.5809 |
| TG373 | 1.1286 | 180.8 | 0.43 | 179.6714 |
| TG4233 | 306.6894 | 126.2 | 0.05 | 179.5106 |
| TG5289 | 277.0299 | 97.5 | 0.03 | 179.5299 |
| TG5911 | 286.8179 | 105.5 | 0.31 | 178.6821 |
| TG6267 | 244.3546 | 63.7 | 0.15 | 179.3454 |
| TG9856 | 215.5961 | 34.7 | 0.08 | 179.1039 |